\documentclass[journal]{IEEEtran}
\usepackage{cite}
\usepackage[cmex10]{amsmath}
\usepackage[tight,footnotesize]{subfigure}
\usepackage{float}
\usepackage{stfloats}
\usepackage{url}
\usepackage{graphicx}
\usepackage{multirow}
\usepackage[colorlinks,
            linkcolor=red,
            anchorcolor=blue,
            citecolor=green
            ]{hyperref}
\usepackage{booktabs}
\usepackage[table]{xcolor}

\begin{document}
%
\title{Learn to Evaluate Image Perceptual Quality Blindly from Statistics of Self-similarity}
%
%
%

\author{Wufeng~Xue,~\IEEEmembership{Student Member,~IEEE,}
        Xuanqin~Mou,~\IEEEmembership{Member,~IEEE,}
        and~Lei~Zhang,~\IEEEmembership{Senior~Member,~IEEE}
\thanks{This work was supported in part by National Natural Science Foundation of China under Grant 61172163, Grant 90920003,
and Grant 61271294, and in part by HK RGC GRF grant (under no. PolyU 5313/13E).}
\thanks{W. Xue and X. Mou are with the Institute of Image Processing and Pattern Recognition, Xi'an Jiaotong University, Xi'an, China. (Email: xwolfs@hotmail.com, xqmou@mail.xjtu.edu.cn). X. Mou is also with the Beijing Center for Mathematics and Information Interdisciplinary Sciences (BCMIIS), Beijing, China.}
\thanks{L. Zhang is with the Department of Computing, The Hong Kong Polytechnic University, Hong Kong. (Email: cslzhang@comp.polyu.edu.hk).}
}

%
%

\markboth{IEEE Transaction on Multimedia,~Vol.~X, No.~X, XX~20XX}
{Xue \MakeLowercase{\textit{et al.}}: Learn to Evaluate Image Perceptual Quality from SOS}
%



\maketitle

\begin{abstract}
Among the various image quality assessment (IQA) tasks, blind IQA (BIQA) is particularly challenging due to the absence of knowledge about the reference image and distortion type. Features based on natural scene statistics (NSS) have been successfully used in BIQA, while the quality relevance of the feature plays an essential role to the quality prediction performance. Motivated by the fact that the early processing stage in human visual system aims to remove the signal redundancies for efficient visual coding, we propose a simple but very effective BIQA method by computing the statistics of self-similarity (SOS) in an image. Specifically, we calculate the inter-scale similarity and intra-scale similarity of the distorted image, extract the SOS features from these similarities, and learn a regression model to map the SOS features to the subjective quality score. Extensive experiments demonstrate very competitive quality prediction performance and generalization ability of the proposed SOS based BIQA method.
\end{abstract}

\begin{IEEEkeywords}
Blind image quality assessment, natural scene statistics, self-similarity, image redundancy.
\end{IEEEkeywords}

%
\IEEEpeerreviewmaketitle

\section{Introduction}

\IEEEPARstart{I}{mage} quality assessment (IQA) aims to measure to what extent the observer is satisfied with the perceptual quality of a given image. IQA has become increasingly important due to its versatile utilities, including image quality monitoring, parameter tuning of image processing algorithms, and acting as yardstick for image processing system performance evaluation. With the proliferation of applications of high speed networks and portable multimedia devices, the demanding of reliable and efficient IQA algorithms is getting higher.

In the past decade, a variety of IQA methods have been proposed, which can be generally classified into three categories according to the available information of original reference image~\cite{wang2006modern_1}: full reference (FR), reduced reference (RR) and no reference (NR). The FR methods have high prediction accuracy~\cite{zhang2011nser_16,larson2010most_21,xue2014gradient_30,SongnanFR,zhang2011fsim} because of the availability of the pristine reference image. In RR methods, a brief description of the reference image is available, for example, features based on natural scene statistics (NSS). By matching the statistics between the reference image and the distorted image, RR methods can also lead to good accuracy of quality prediction~\cite{li2009reduced,xue2010weibull,mou2012reduced,LinMA_DCT_RR,VIF_RR}. However, both FR and RR methods are hard to use in many practical applications, where the reference image information is completely inaccessible. Therefore, it is highly demanding to develop NR methods to predict the quality of a distorted image without prior information.

The NR methods can be categorized into distortion-specific (DS) methods and non-distortion-specific (NDS) ones. DS methods assume that the image degradation procedure is known, and descriptors that are capable of capturing the artifacts are employed to measure the quality. A review of DS methods can be found in~\cite{shahid2014no_2}. In NDS methods, the distortion procedure is unknown, which is the case in most practical applications. Usually, this class of NR methods are also called blind IQA (BIQA) methods.

\begin{figure*}[!t]
  \centering
  \includegraphics[width=0.90\linewidth]{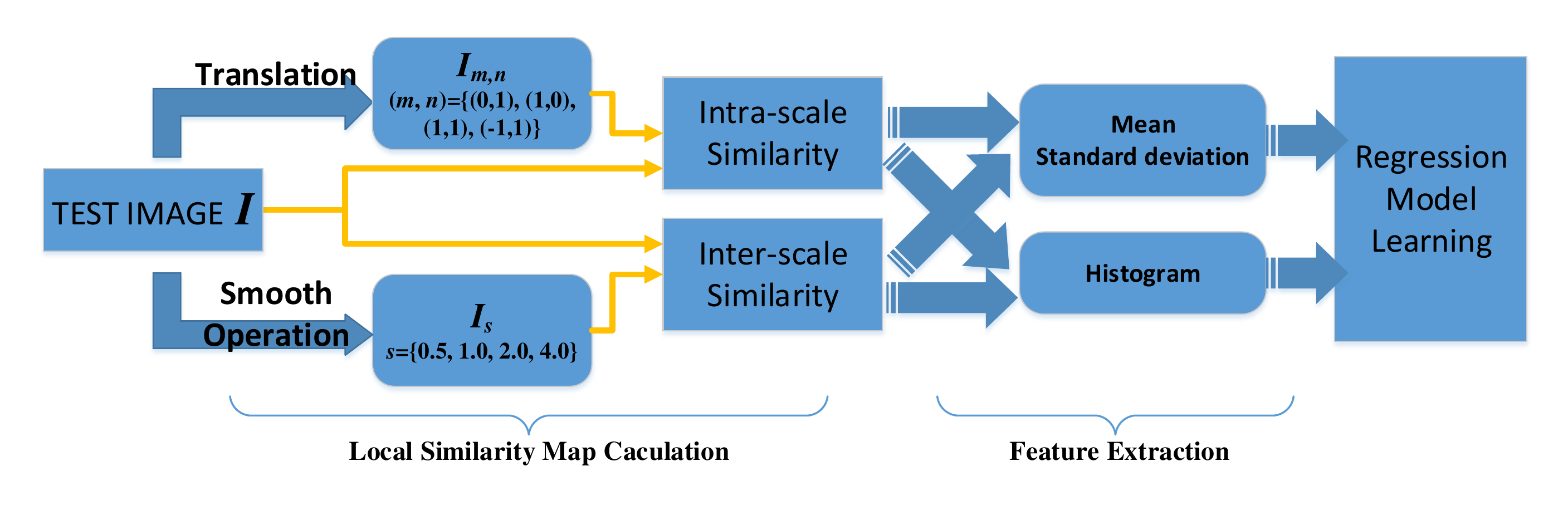}
  \vspace{-5mm}
  \caption{Flowchart of the proposed BIQA method by statistics of self-similarity (SOS).}
  \label{fig_1}

\end{figure*}

Existing solutions to BIQA have achieved good performance with the help of machine learning methods such as support vector regression and neural network. A survey about recent BIQA methods can be found in~\cite{manap2015non_3}. These methods differ from each other mainly on how the quality aware features are extracted. It is widely accepted that natural images are highly sparse in the high dimensional space. Once a natural image is distorted, its characteristics will accordingly deviate from that of the original image. This provides the underlying motivation of most BIQA methods. To capture the quality aware representation, usually images are subjected to decompositions of multiple frequencies and orientations by using wavelet~\cite{moorthy2010biqi_9,tang2011lbiq_10,Moorthy2011blind_23}, contourlet~\cite{lu2010no_11} and discrete cosine transform (DCT)~\cite{saad2010blinddct_12,Saad2011model_13}, etc. Compared with the pixel based representation, redundancies among the coefficients in these transformed domain are largely reduced. The resulting transform coefficients follow a high kurtosis, heavy tailed distribution. Distortions presented in an image will lead to deviation of this distribution, which can be used to predict image's quality.

To further reduce the redundancy of images for a more effective (in the viewpoint of encoding) representation, contrast normalization is introduced to multi-scale and multi-orientation image decomposition~\cite{teo1994perceptual_4}. It is shown~\cite{ruderman1994statistics_5} that the mean-subtracted contrast normalized (MSCN) coefficients are decorrelated and follow Gaussian distribution~\cite{ruderman1994statistics_5}. This finding can be used to model the contrast masking effect in early human vision. Mittal et al.~\cite{mittal2012no_6} parameterized MSCN coefficients with a Generalized Gaussian Distribution (GGD) and the pairwise product of MSCN coefficients with an asymmetric GGD (AGGD). The resulting method, called BRISQUE, obtains state-of-the-art BIQA performance. Inspired by this contrast normalization, Xue et al.~\cite{xue2014blind_7} proposed a jointly adaptive normalization (JAN) scheme to reduce the redundancy in domains of Laplacian of Gaussian (LOG) response and gradient magnitude (GM). After the JAN operation, the statistics of LOG and GM become more similar among natural images of different contents while becoming more different from unnatural images. The proposed model M3 in~\cite{xue2014blind_7} shows better performance than BRISQUE on two benchmark databases.

Most existing BIQA methods~\cite{moorthy2010biqi_9,tang2011lbiq_10,Moorthy2011blind_23,lu2010no_11,saad2010blinddct_12,Saad2011model_13} calculate the image statistics in a transformed domain where the image redundancies are much reduced. Contrary to these methods, we find that measuring image redundancy directly in the pixel based spatial domain can lead to an efficient BIQA method with promising performance. Human visual system has evolved to economically describe natural images by efficient redundancy reduction~\cite{attneave1954some_8}. The redundancy in an image can be reflected by the predictability of a pixel's intensity by its neighboring pixels~\cite{kersten1987predictability}. Natural images generally have high spatial correlation and multi-scale correlation; that is, a natural image looks similar to its translated, zoom in or zoom out versions. Therefore, we measure the image redundancy by computing the image intra-scale and inter-scale self-similarities, and propose a BIQA method called Statistics of Self-similarity (SOS). It is worthwhile to note that SOS is very different from the previous work M3 in~\cite{xue2014blind_7}. First, SOS aims to describe the degree of redundancy of an image, while M3 aims to capture the local contrast. Second, the SOS features are based on the distributions of similarity maps, while in M3 the GM and LOG features are used and jointly normalized to obtain a more robust feature representation. At last, SOS provides a general framework of BIQA and any similarity function can be employed in it.

The rest of this paper is organized as follows. Section~\ref{II} presents in details the proposed SOS computation framework, and demonstrates the high relevance of SOS based features with image quality. Section~\ref{III} gives experimental settings. Extensive experimental results and analysis are presented in Section~\ref{IV}. Section~\ref{V} concludes the paper.

\section{STATISTICS OF SELF-SIMILARITY FOR BIQA}\label{II}
The flowchart of the proposed SOS based BIQA method is illustrated in Fig.~\ref{fig_1}. It consists of the following main steps: local self-similarity map calculation, SOS feature extraction, and regression model learning, which are described in detail as follows.

\begin{figure*}[!tb]
\centering
\subfigure[Reference image.]{
\label{fig:2a}
\includegraphics[width=0.30\linewidth]{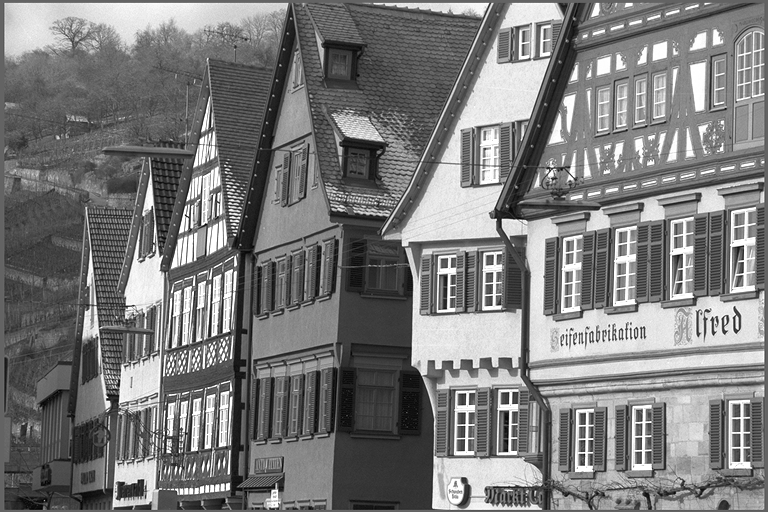}
}
\hspace{0.1in}
\subfigure[Inter-scale LSM 1 for (a)]{
\label{fig:2b}
\includegraphics[width=0.30\linewidth]{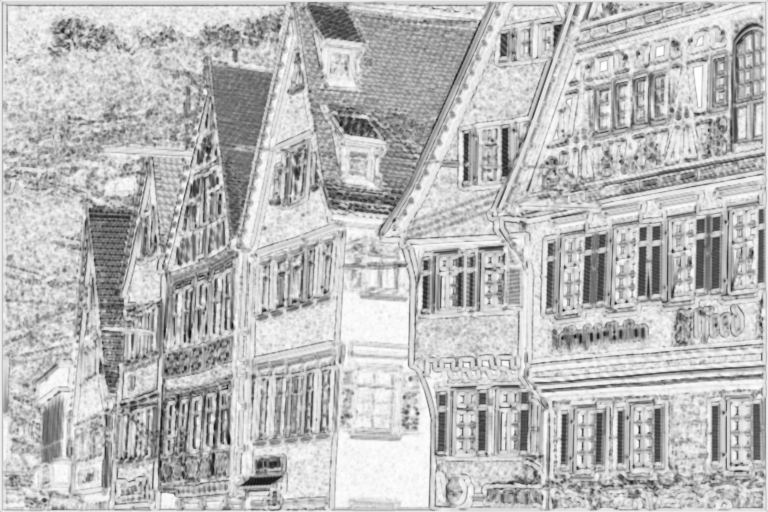}
}
\hspace{0.1in}
\subfigure[Inter-scale LSM 2 for (a)]{
\label{fig:2c}
\includegraphics[width=0.30\linewidth]{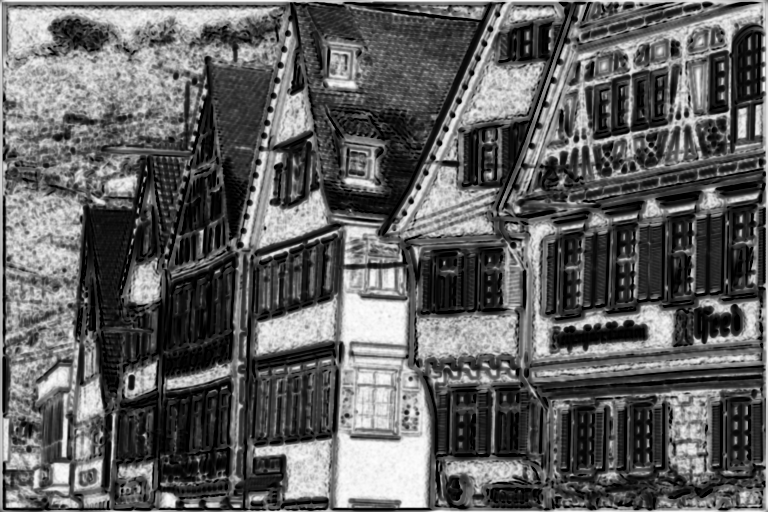}
}

\hspace{2.4in}
\subfigure[Intra-scale LSM 1 for (a)]{
\label{fig:2d}
\includegraphics[width=0.30\linewidth]{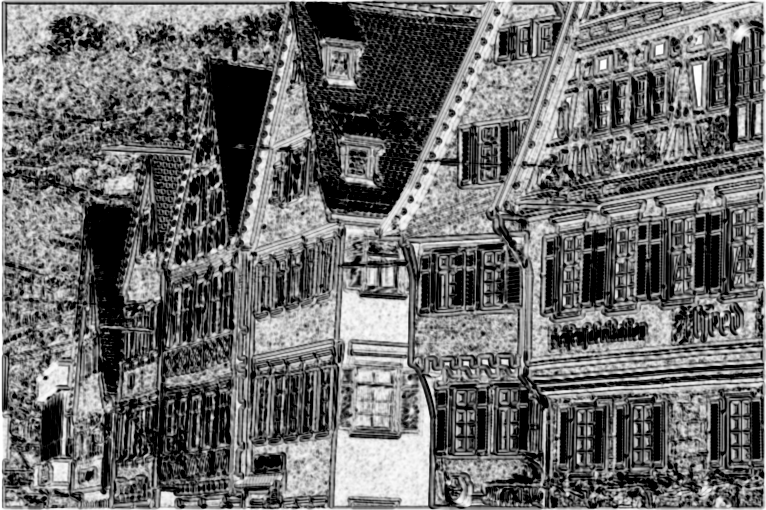}
}
\hspace{0.1in}
\subfigure[Intra-scale LSM 2 for (a)]{
\label{fig:2e}
\includegraphics[width=0.30\linewidth]{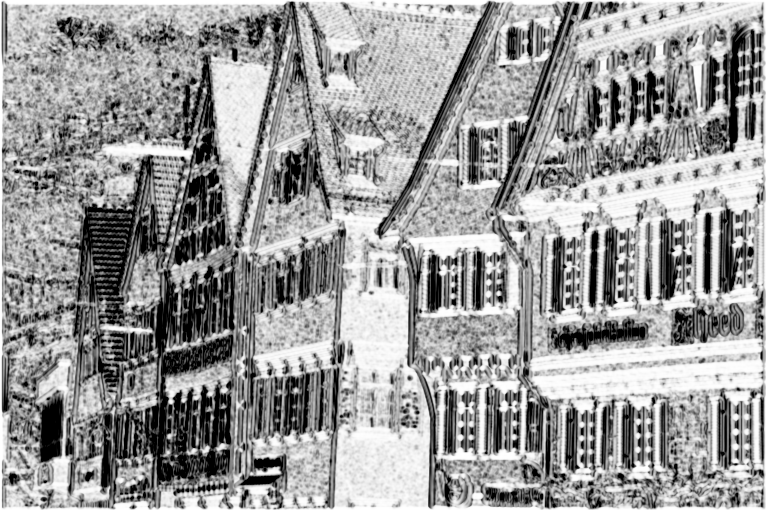}
}
\subfigure[Distorted image]{
\label{fig:2f}
\includegraphics[width=0.30\linewidth]{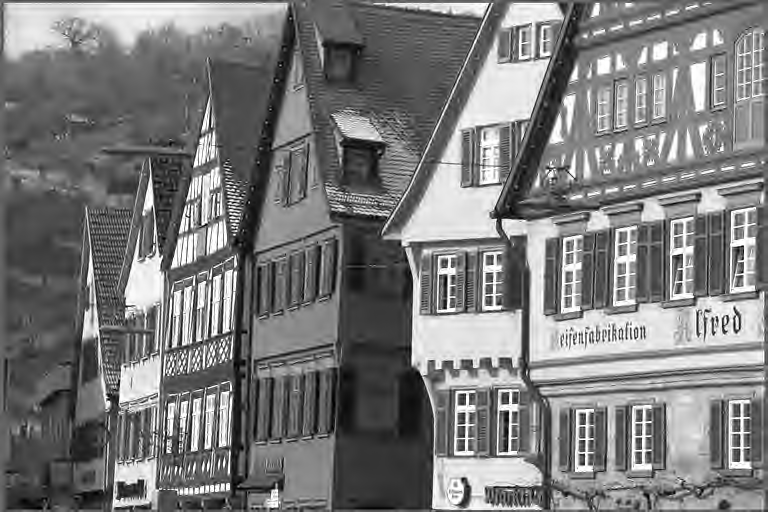}
}
\hspace{0.1in}
\subfigure[Inter-scale LSM 1 for (f)]{
\label{fig:2g}
\includegraphics[width=0.30\linewidth]{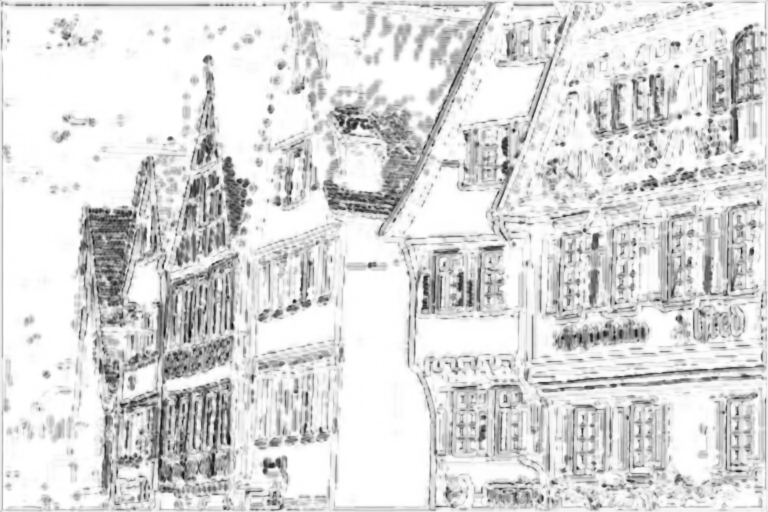}
}
\hspace{0.1in}
\subfigure[Inter-scale LSM 2 for (f)]{
\label{fig:2h}
\includegraphics[width=0.30\linewidth]{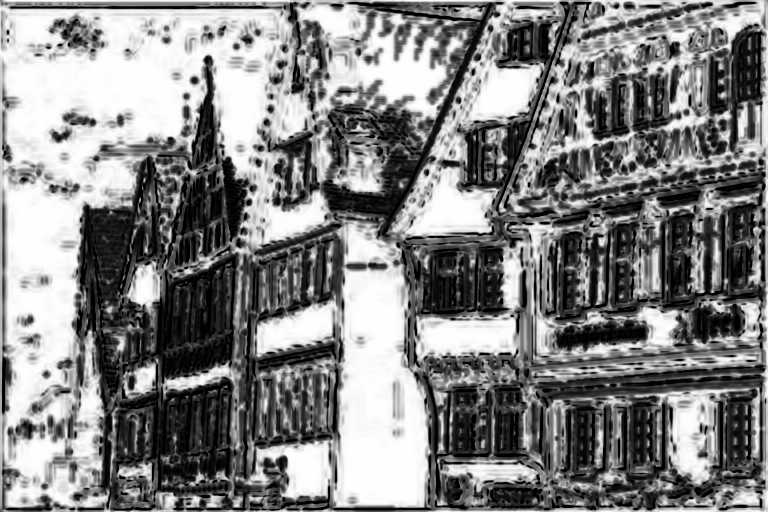}
}

\hspace{2.4in}
\subfigure[Intra-scale LSM 1 for (f)]{
\label{fig:2i}
\includegraphics[width=0.30\linewidth]{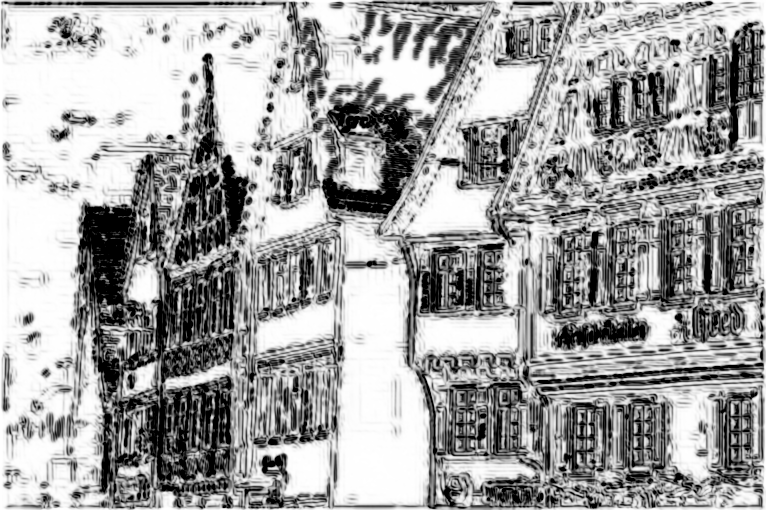}
}
\hspace{0.1in}
\subfigure[Intra-scale LSM 2 for (f)]{
\label{fig:2j}
\includegraphics[width=0.30\linewidth]{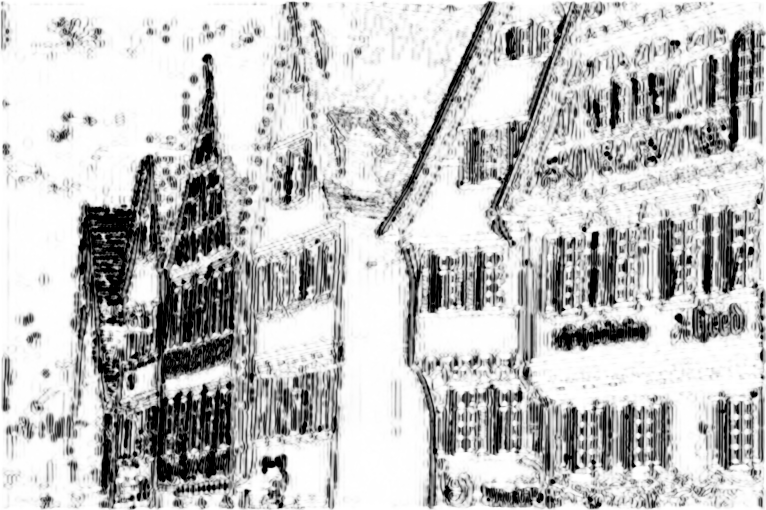}
}

\caption{The LSMs of a reference image and its JPEG2000 compressed image.}
\label{fig_2} 
\end{figure*}

\subsection{Calculation of Local Self-similarity Map}\label{II.1}
For an input image \textbf{I}, the redundancy can be described by its self-similarity. In particular, the image self-similarity can be measured from two aspects. The first one is intra-scale self-similarity. Due to the spatial correlation, \textbf{I} will be similar to its translated versions, denoted by $\textbf{I}_{m,n}$, where \textit{m} and \textit{n} are the translations along vertical and horizontal directions. Refer to Fig.~\ref{fig_1}, in this paper we employ four translated versions of \textbf{I} by setting $(m, n)={(0,1), (1,0), (1,1), (-1,1)}$.

Apart from intra-scale self-similarity, natural images also exhibit inter-scale self-similarity. It is well known that natural images have the property of scale invariance, i.e., an image usually looks similar to its scaled versions. Considering the fact that the scale space of human visual system can be well approximated by Gaussian filtering~\cite{dam2003front_14}, we produce a series of smoothed versions of \textbf{I} by:
\begin{equation}\label{EQ_smooth}
\textbf{I}_s(x,y)=\sum_{(d_x,d_y)\in support(\textbf{h}_s)}{\textbf{I}(x-d_x,y-d_y)\textbf{h}_s(d_x,d_y)},
\end{equation}
where \emph{x}, \emph{y} are the spatial location, and
\begin{equation}\label{EQ_gaussian}
    \textbf{h}_s(d_x,d_y)=\frac{1}{2\pi s^2}exp{(-\frac{d_x^2+d_y^2}{2s^2})}
\end{equation}
is the 2D Gaussian filter with scale \emph{s}. Refer to Fig.~\ref{fig_1}, we compute four smoothed versions of \textbf{I} with $s=\{0.5, 1.0, 2.0, 4.0\}$.

Denote by \textbf{R} any one of the four translated images $\textbf{I}_{m,n}$ and the four smoothed images $\textbf{I}_s$. Both the intra-scale and inter-scale self-similarity can be calculated by computing the similarity between \textbf{I} and \textbf{R} in any local region, leading to a Local Similarity Map (LSM). Intuitively, the similarity functions used in many existing FR IQA methods to compute the local quality map (LQM) can all be used to compute this LSM. In this paper, we adopt the similarity functions in two representative FR-IQA methods, i.e., Structural SIMilarity (SSIM)~\cite{wang2004ssim_15} and ratio of non-shift edge (rNSE)~\cite{zhang2011nser_16,xue2011rnse_17}.

\begin{figure*}[!tb]
\centering
\subfigure{
\label{fig:3a}
\includegraphics[width=0.40\linewidth]{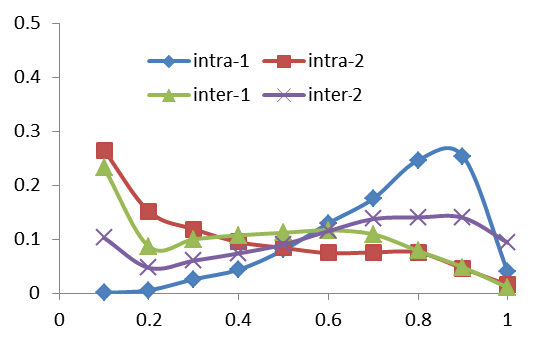}
}
\hspace{0.1in}
\subfigure{
\label{fig:3b}
\includegraphics[width=0.40\linewidth]{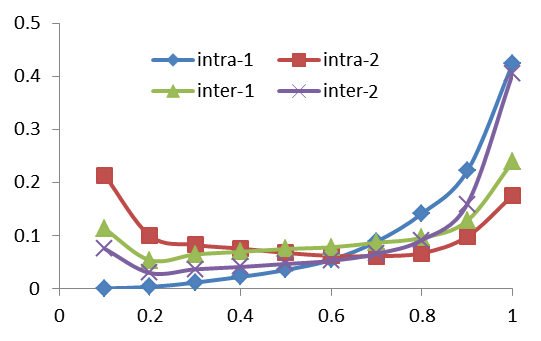}
}

\caption{The distributions of the LSMs for Fig.~\ref{fig_2}(a) (left) and Fig.~\ref{fig_2}(f) (right).}
\label{fig_3} 
\vspace{-2mm}
\end{figure*}

\begin{table*}
  \centering
  \caption{The SOS features considered in the proposed method.}
  \begin{tabular}{*{3}{c}}
  \toprule
  Feature &	Description	& Dimension (for each of the 8 LSMs)\\
  \midrule
    $\mu$&     	Mean of the elements in an LSM&	1 (8 in total)\\
    \textit{d}&	Standard deviation of the elements in an LSM&	1 (8 in total)\\
    \textbf{\textit{h}}&	Histogram of a quantized (10 bins) LSM&	10 (80 in total)\\
  \bottomrule
  \end{tabular}
  \vspace{-3mm}
  \label{Table_1}
\end{table*}

SSIM is a benchmark FR IQA method. In SSIM, the local similarity at each location $(x, y)$ is calculated by~\cite{wang2004ssim_15}:
\begin{equation}\label{EQ_SSIM}
    LSM_{x,y}(\textbf{I},\textbf{R})=(L_{x,y}(\textbf{I},\textbf{R}))^\alpha
    (C_{x,y}(\textbf{I},\textbf{R}))^\beta(S_{x,y}(\textbf{I},\textbf{R}))^\gamma
\vspace{-2mm}
\end{equation}
where constants $\alpha$, $\beta$ and $\gamma$ mediate the relative importance of the three components. \emph{L}, \emph{C} and \emph{S} measure the similarities of luminance, contrast and structure between \textbf{I} and \textbf{R}:
\begin{equation}\label{EQ_L}
    L_{x,y}(\textbf{I},\textbf{R})=\frac{2\mu_\textbf{I}(x,y)\mu_\textbf{R}(x,y)+c_1}
    {\mu_\textbf{I}^2(x,y)+\mu_\textbf{R}^2(x,y)+c_1},
\end{equation}
\begin{equation}\label{EQ_C}
    C_{x,y}(\textbf{I},\textbf{R})=\frac{2\sigma_\textbf{I}(x,y)\sigma_\textbf{R}(x,y)+c_2}
    {\sigma_\textbf{I}^2(x,y)+\sigma_\textbf{R}^2(x,y)+c_2},
\end{equation}
\begin{equation}\label{EQ_S}
    S_{x,y}(\textbf{I},\textbf{R})=\frac{\sigma_{\textbf{I},\textbf{R}}(x,y)+c_3}
    {\sigma_\textbf{I}(x,y)\sigma_\textbf{R}(x,y)+c_3},
\end{equation}
where $\mu_\textbf{I}$ and $\mu_\textbf{R}$ are the local means of \textbf{I} and \textbf{R}; $\sigma_\textbf{I}$ and $\sigma_\textbf{R}$ are the local standard deviations of \textbf{I} and \textbf{R}; and $\sigma_{\textbf{I},\textbf{R}}$ is the local covariance between \textbf{I} and \textbf{R}. All these computations are applied using a local Gaussian window with a specified scale parameter as the weighting factor. $c_1$, $c_2$ and $c_3$ are small constants to avoid the denominator being zero. In this work, we follow~\cite{wang2004ssim_15} for the configuration of the parameters $\alpha$, $\beta$, $\gamma$, $c_1$, $c_2$ and $c_3$.

rNSE~\cite{zhang2011nser_16,xue2011rnse_17} is a recently proposed FR-IQA method, which measures the image quality by computing the ratio of the number of non-shift edges after distortion to the number of original edges. The rNSE index is computed as:
\begin{equation}\label{EQ_rNSE}
    rNSE(\textbf{E}_\textbf{A},\textbf{E}_\textbf{D})=\frac{|{\textbf{E}_\textbf{A}}\cap {\textbf{E}_\textbf{D}}|}{|{\textbf{E}_\textbf{A}}|},
\end{equation}
where $\textbf{E}_\textbf{D}$ and $\textbf{E}_\textbf{A}$ are the sets of edges of distorted image \textbf{D} and reference image \textbf{A}, respectively. The computation of the edge set is based on the zero-crossings detection of the Laplacian of Gaussian (LOG) response~\cite{zhang2011nser_16,xue2011rnse_17}. "$\cap$" denotes the intersection of sets $\textbf{E}_\textbf{D}$ and $\textbf{E}_\textbf{A}$, i.e., non-shift edge points between \textbf{A} and \textbf{D}; "$|\bullet|$" counts the number of edges in the set. Clearly, rNSE is a ratio between 0 and 1. Since in the context of BIQA, the reference image \textbf{A} is not available, we modify the rNSE index as follows to calculate the desired LSM:
\begin{equation}\label{Eq_rNSE2}
    LSM_{x,y}(\textbf{I},\textbf{R})=\frac{2|\textbf{E}_\textbf{I}^{(x,y)}\cap \textbf{E}_\textbf{R}^{(x,y)}|}{|\textbf{E}_\textbf{I}^{(x,y)}|+
    |\textbf{E}_\textbf{R}^{(x,y)}|},
\end{equation}
where $\textbf{E}_\textbf{I}^{(x,y)}$ and $\textbf{E}_\textbf{R}^{(x,y)}$ are respectively the sets of edge points of \textbf{I} and \textbf{R} in a local square window centered at $(x, y)$.

\begin{figure*}[!tb]
  \centering
  \includegraphics[width=0.90\linewidth]{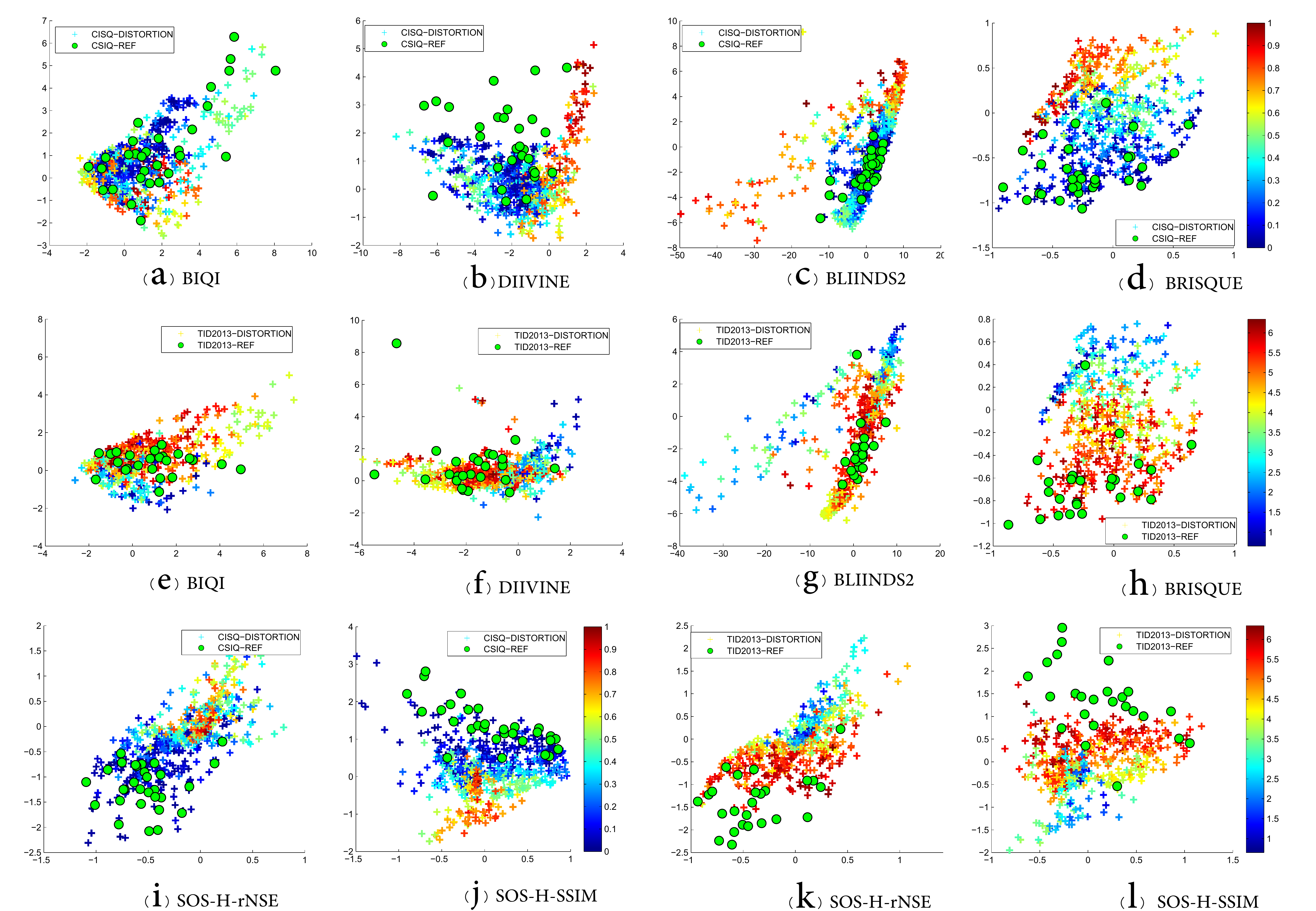}\\
  \caption{2-D visulization of BIQA features. For each BIQA feature, a linear transform is learned from the LIVE database. Then the transform is applied to the coppresponing features in databases of CSIQ and TID2013. The pristine images are represeted as green dots while the distorted images are represented as "+". The color of "+" encodes the subjective score, i.e., DMOS for CSIQ and MOS for TID2013. The used BIQA features include BIQI, DIIVINE, BLIINDS2, BRISQUE, SOS-H-rNSE and SOS-H-SSIM.}
  \label{fig_nca}
\end{figure*}

With either Eq.~\ref{EQ_SSIM} or Eq.~\ref{Eq_rNSE2}, we could calculate eight LSMs of \textbf{I}. In Fig.~\ref{fig_2}, we show the LSMs of a reference image and its distorted image (JPEG2000 compressed). SSIM is used as the similarity function. Two intra-scale LSMs and two inter-scale LSMs are shown. Note that the distortion in Fig.~\ref{fig_2}(f) is moderate and the artifacts generated by compression are nearly invisible. However, from the LSMs, we can easily tell the difference between the reference image and the distorted one. Those LSMs reflect the local correlation of the distorted image along different orientations (for intra-scale self-similarity) and in scale space (for inter-scale self-similarity), implying that the image perceptual quality can be well inferred from them.

\subsection{SOS Features}
From the 8 SOS based LSMs computed above, features can be extracted to predict the image quality. Clearly, the most important statistics of the LSMs are their mean and standard deviation, which are computed as follows:
\begin{eqnarray}
  \mu &=& \frac{1}{N}\sum_x\sum_y LSM_{x,y}, \\
  d &=& \sqrt{\frac{1}{N}\sum_x\sum_y (LSM_{x,y}-\mu)^2},
\end{eqnarray}
where \textit{N} is the number of elements in the LSM. We call the method that uses mean and standard deviation as SOS-MD. One advantage SOS-MD is its low feature dimensionality (16 in total for all the 8 LSMs). Note that the pair $(\mu, d)$ can completely characterize the statistical information of one LSM if its elements follow a Gaussian distribution. However, in practice the distribution of the LSMs are far from Gaussian (please refer to Fig.~\ref{fig_3} for example distributions of the LSMs), and using only the mean and standard deviation cannot accurately describe them. Therefore, we quantify the LSMs into several levels, and use their normalized histograms as the SOS features. We call this method as SOS-H. Since both the SSIM and rNSE indices range between 0 and 1, we quantify each LSM into 10 bins (with step length 0.1), resulting in a 10 dimensional histogram \emph{\textbf{h}} for each LSM.

In TABLE~\ref{Table_1} we list the three types of SOS features: $\mu, d$ and \emph{\textbf{h}}. By using each type of features or the combination of them, we could learn a regression model for BIQA.

\subsection{Regression Model Learning}
With the SOS feature vector, denoted by \emph{\textbf{f}}, of an image, a regression function \emph{\textbf{F}} could be learned to map \emph{\textbf{f}} to the image subjective quality score \emph{q}, i.e., $q\approx \textbf{\emph{F}}(f)$. To this end, we need a set of training images, whose subjective quality scores are available. Such a training dataset can be extracted from the existing IQA databases. We can construct a training set of \textit{k} images with their feature vectors and subjective scores: $\{(\textbf{\textit{f}}_1,q_1), (\textbf{\textit{f}}_2,q_2),бн, (\textbf{\textit{f}}_k,q_k)\}$. Machine learning tools, such as support vector regression (SVR), neuron network, random forest, can be used to learn the mapping \textbf{\textit{F}}. In this work, we adopt the SVR with a radial basis function kernel~\cite{smola2004svrtutorial_18}. The readers may refer to~\cite{smola2004svrtutorial_18} for the details of SVR and its implementation. Once the regression model \emph{\textbf{F}} is learned, we can use it to estimate the perceptual quality of any input image.

\subsection{Comparison with Other NSS based Features}
As we discussed in the Section I, the differences between the SOS based features and the NSS features used in previous BIQA methods lie in two folds. 1) First, instead of transforming the image into another redundancy-reduced domain, we use the translated and smoothed versions of the image in the spatial domain for feature extraction. 2) Second, the statistics of self-similarity maps are used as the quality aware features. To illustrate the power of SOS features in BIQA, we use the neighboring components analysis (NCA)~\cite{roweis17neighbourhood_19} to transform the high dimensional BIQA features into 2 dimensional (2-D) points, and then plot the scatter of these 2-D points to reveal the essential structure of the data.

We first learn a projection matrix via NCA on the LIVE database~\cite{sheikh2007live_20} for each BIQA feature, then apply this matrix to features on databases of CSIQ~\cite{larson2010most_21} and TID2013~\cite{tid2013_22}. The scatter plots of the resulting 2-D points are shown in Fig.~\ref{fig_nca}. In each plot, the reference images are represented as green dots, while the distorted images as "+". The color of "+" encodes the subjective score of each distorted image. The used BIQA features include BIQI~\cite{moorthy2010biqi_9}, DIIVINE~\cite{Moorthy2011blind_23}, BLIINDS-II~\cite{Saad2011model_13}, BRISQUE~\cite{mittal2012no_6}, SOS-H-rNSE and SOS-H-SSIM. The first row shows the scatter plots of the 2-D points which are obtained when the LIVE-learned projection matrix is applied to the CSIQ database, while the second row shows the results on the TID2013 database. The third row in Fig.~\ref{fig_nca} shows the corresponding plots for SOS-H-rNSE and SOS-H-SSIM.

From these plots, we can draw the following conclusions. 1) The distributions of the 2-D points for BRISQUE, SOS-H-rNSE and SOS-H-SSIM show obvious quality relevance. The points for reference images are close or overlapped with points of slightly distorted images and far from the heavily distorted images. The intermediately distorted images are by and large ordered according to their quality. 2) In the distributions for BIQI, DIIVINE and BLIINDS-II, no or weak quality relevance can be observed. Distorted images are not sequentially located according to their quality. These observations reveal the advantages of BRISQUE and the proposed SOS-H methods over the other methods.

\begin{table*}
    \newcommand{\tabincell}[2]{\begin{tabular}{@{}#1@{}}#2\end{tabular}}
  \centering
  \caption{\textsc{Performance comparison of SOS-MD and SOS-H on the three databases. Results for the overall database and for each individual distortion are listed.}}
  \begin{tabular}{cc*{4}{|ccc}}
  \toprule
  \multicolumn{2}{c}{\multirow{2}*{\tabincell{c}{Database\\Distortion type}}}
                    &\multicolumn{3}{|c}{SOS-MD-rNSE}& \multicolumn{3}{|c}{SOS-MD-SSIM}& \multicolumn{3}{|c}{SOS-H-rNSE}
                   & \multicolumn{3}{|c}{SOS-H-SSIM}\\

                   & &PCC	&RMSE	&SRC	&PCC	&RMSE	&SRC	&PCC	&RMSE	&SRC	 &PCC	&RMSE	&SRC\\
  \midrule
  \multirow{6}*{LIVE}
    &ALL	&\textbf{0.918}	&\textbf{10.764}	&\textbf{0.911}	&0.891	&12.373	&0.871	 &\textbf{0.946}	&\textbf{8.837}	 &\textbf{0.943}	 &0.926	&10.234	&0.921\\
    &JP2K	&\textbf{0.923}	&\textbf{9.516}	&\textbf{0.904}	&0.849	&13.119	&0.825	 &\textbf{0.950}	&\textbf{7.900}	 &\textbf{0.933}	 &0.923	 &9.656	&0.906\\
    &JPEG	&\textbf{0.963}	&\textbf{8.564}	&\textbf{0.945}	&0.937	&11.088	&0.911	 &\textbf{0.974}	&\textbf{7.159}	 &\textbf{0.958}	 &0.962	 &8.672	&0.942\\
    &WN	&0.978	&5.811	&0.967	&\textbf{0.983}	&\textbf{5.194}	&\textbf{0.973}	 &\textbf{0.989}	&\textbf{4.121}	&\textbf{0.980}	 &0.981	 &5.486	&0.971\\
    &GB	&0.912	&7.673	&0.877	&\textbf{0.925}	&\textbf{6.976}	&\textbf{0.891}	 &\textbf{0.923}	&\textbf{7.018}	&\textbf{0.903}	 &0.917	 &7.372	&0.875\\
    &FF	&\textbf{0.884}	&\textbf{13.113}	&\textbf{0.844}	&0.855	&14.296	&0.738	 &\textbf{0.918}	&\textbf{11.094}	 &\textbf{0.890}	 &0.903	&12.109	&0.868\\
  \midrule
  \multirow{5}*{CSIQ}
    &ALL	&\textbf{0.926}	&\textbf{0.104}	&\textbf{0.903}	&0.890	&0.129	&0.863	 &\textbf{0.933}	&\textbf{0.101}	 &\textbf{0.910}	 &0.926	 &0.106	&0.901\\
    &WN	&0.946	&0.054	&\textbf{0.937}	&\textbf{0.954}	&\textbf{0.050}	&0.933	&0.938	 &0.059	&\textbf{0.923}	&\textbf{0.941}	 &\textbf{0.056}	&0.920\\
    &JPEG	&\textbf{0.967}	&\textbf{0.078}	&\textbf{0.922}	&0.927	&0.114	&0.885	 &\textbf{0.961}	&\textbf{0.084}	&0.906	 &0.957	 &0.088	 &\textbf{0.915}\\
    &JP2K	&\textbf{0.935}	&\textbf{0.110}	&\textbf{0.912}	&0.875	&0.152	&0.842	 &\textbf{0.941}	&\textbf{0.106}	 &\textbf{0.915}	 &0.936	 &0.111	&0.907\\
    &GB	&\textbf{0.917}	&\textbf{0.112}	&\textbf{0.879}	&0.906	&0.119	&0.851	&0.929	 &0.103	&0.902	&\textbf{0.935}	 &\textbf{0.099}	 &\textbf{0.905}\\
  \midrule
  \multirow{5}*{TID2008}
    &ALL	&\textbf{0.931}	&\textbf{0.514}	&\textbf{0.927}	&0.902	&0.603	&0.871	&0.937	 &0.488	&\textbf{0.928}	&\textbf{0.938}	 &0.488	 &0.920\\
    &WN	&0.905	&0.302	&0.886	&\textbf{0.924}	&\textbf{0.270}	&\textbf{0.906}	 &\textbf{0.940}	&\textbf{0.243}	&\textbf{0.923}	 &0.934	 &0.252	&0.914\\
    &GB	&\textbf{0.925}	&\textbf{0.466}	&\textbf{0.912}	&0.894	&0.556	&0.876	&0.923	 &0.475	&0.913	&0.923	&\textbf{0.469}	 &\textbf{0.914}\\
    &JPEG	&\textbf{0.971}	&\textbf{0.358}	&\textbf{0.923}	&0.928	&0.547	&0.842	 &\textbf{0.969}	&\textbf{0.367}	 &\textbf{0.906}	 &0.965	 &0.394	&0.901\\
    &JP2K	&\textbf{0.945}	&\textbf{0.556}	&\textbf{0.929}	&0.918	&0.688	&0.872	 &\textbf{0.950}	&\textbf{0.535}	 &\textbf{0.931}	 &0.948	 &0.540	&0.923\\

  \bottomrule
  \end{tabular}
  \label{Table_2}
\end{table*}

\section{EXPERIMENTS CONFIGURATION}\label{III}
\subsection{Image Databases and Evaluation Criteria}
We evaluate the performance of the proposed SOS based BIQA methods in terms of their ability to predict the subjective score of distorted images. Three publicly available large-scale databases are employed for this evaluation.

\textbf{The LIVE database}~\cite{sheikh2007live_20}: A total of 779 distorted images are generated by applying 5 distortion operations at $4\sim5$ levels to 29 pristine images. The distortions include: JPEG2000 compression (JP2K), JPEG compression, white noise (WN), Gaussian blurring (GB) and simulated fast fading Rayleigh channel (FF).

\textbf{The CSIQ database}~\cite{larson2010most_21}: A total of 866 distorted images are generated by applying 5 distortion operations at $3\sim5$ levels to 30 pristine images. The distortions include: JPEG, JP2K, additive pink noise, WN, GB and global contrast decrements.

\textbf{The TID2013 database}~\cite{tid2013_22}: A total of 3000 distorted images are generated by applying 24 distortion operations at 5 levels to 25 pristine images. The distortions in TID2013 reflect a broad range of image impairments, such as edge smoothing, block artifacts, additive and multiplicative noise, chromatic aberrations, denoising and contrast change, etc. Details of the distortions can be found in~\cite{tid2013_22}.

The ground truth quality of each image is given by the subjective score, i.e., (Difference) Mean Opinion Score (DMOS/MOS). To evaluate the performance of BIQA methods, three indexes are usually computed by using the subjective scores and the model-predicted scores: the Spearman rank order correlation coefficient (SRC), the Pearson correlation coefficient (PCC) after a logistic regression~\cite{video2003final_24}, and the root mean squared error (RMSE) between the subjective score and the predicted score after the regression. Note that this logistic regression accounts for the different range of the objective and subjective scores, as well as the nonlinearity of human perception in extreme distortions.

\subsection{Implementation Details and Parameter Setting}\label{III.B}
In the implementation of the proposed BIQA method, the scale of the Gaussian window in SSIM and the scale of the LOG filter in rNSE are both set as 0.5 for computing the intra-scale self-similarity. For the inter-scale self-similarity, we set the scale parameters of SSIM and rNSE to the same value as the four smooth parameters in Eq.~\ref{EQ_gaussian}. With the LSMs available, the SOS features (mean $\mu$, standard deviation \emph{d} and histogram \emph{\textbf{h}}) can be extracted, and then fed into the SVR to train a regression model for quality prediction. We adopt the $\varepsilon$-SVR algorithm with an RBF (radial-basis function) kernel, and the source code is from LibSVM~\cite{chang2011libsvm_25}. The parameters of $\varepsilon$-SVR are tuned by 2D grid search in the logarithm space.

During our experiments, 80\% of the images are employed for training and the rest 20\% for testing. The training and test sets are split according to the reference image to guarantee the independency of the image content in training set and test set. This splitting is repeated for 1,000 times and the median results are used to evaluate the final performance.

\section{Results and Discussions}\label{IV}
\subsection{Performance of SOS-MD and SOS-H}
We first compare the performance of SOS-MD and SOS-H on the three databases. The results are listed in TABLE~\ref{Table_2}. Note that in this experiment we only consider the common distortion types to all the three databases, i.e., JP2K, JPEG, WN, and GB.

From TABLE~\ref{Table_2}, we can observe that both SOS-MD and SOS-H show good performance on the three databases in terms of PCC and SRC. The best results on all the three databases are achieved by the method SOS-H-rNSE, with PCC values 0.946, 0.933, and 0.937 on the three databases, respectively. For every single distortion type, it also demonstrates PCC and SRC values consistently higher than 0.9. Note that due to the scale difference of subjective scores on the three databases, the resulting RMSE values range differently.

As for the two similarity functions, rNSE shows clear advantages over SSIM. We highlight in boldface the better one of rNSE and SSIM in each row. The SOS-MD and SOS-H methods with rNSE as similarity function outperform those with SSIM in most of the distortion types. This may be due to the fact that rNSE emphasizes more on edge structure, which is crucial for human visual perception. Besides, benefiting from the richer information in histograms, SOS-H always exhibits better performance than SOS-MD on the three databases.

\begin{table*}
    \newcommand{\tabincell}[2]{\begin{tabular}{@{}#1@{}}#2\end{tabular}}
  \centering
  \caption{\textsc{Median src of the existing BIQAmethods on live database. The results of PSNR and SSIM are listed for reference. (*the results are computed by using the codes provided by the authors.)}}
  \begin{tabular}{l|*{7}{c}}
  \toprule
    Methods	&Feature domain	&JP2K	&JPEG	&WN	&GB	&FF	&ALL\\
  \midrule
    BIQI~\cite{moorthy2010biqi_9}*	&Wavelet	&0.7849	&0.8801	&0.9157	&0.8367	&0.7023	 &0.8084\\
    GRNN~\cite{li2011blind_26}	&Fourier+Spatial	&0.8156	&0.8721	&0.9794	&0.8331	 &0.7354	&0.8268\\
    LD-GS~\cite{gao2011universal_27}	&Wavelet	&0.8317	&0.8339	&0.9134	&0.8751	 &0.8588	&0.8414\\
    LD-TS~\cite{gao2011universal_27}	&Wavelet	&0.8202	&0.8334	&0.9556	&0.9251	 &0.8863	&0.8833\\
    DIIVINE~\cite{Moorthy2011blind_23}*	&Wavelet	&0.8418	&0.8926	&0.9617	&0.8792	 &0.8202	&0.8816\\
    CBIQ-I~\cite{ye2012no}	&Gabor	&0.912	&0.963	&0.959	&0.918	&0.885	&0.896\\
    BLIINDS-II~\cite{Saad2011model_13}*	&DCT	&\textbf{0.9258}	&0.95	&0.9477	&0.9132	&0.8736	 &0.9302\\
    CBIQ-II~\cite{ye2012no}	&Gabor	&0.919	&\textbf{0.965}	&0.933	&\textbf{0.944}	 &\textbf{0.912}	&0.93\\
    BRISQUE~\cite{mittal2012no_6}*	&Spatial	&0.9175	&\textbf{0.9655}	&\textbf{0.9789}	&\textbf{0.9479}	 &0.8854	 &\textbf{0.943}\\
    M3~\cite{xue2014blind_7}	&Gradient magnitude+LOG	&\textbf{0.9283}	 &\textbf{0.9659}	&\textbf{0.9853}	&\textbf{0.9359}	 &\textbf{0.9008}	 &\textbf{0.9511}\\
    SOS-H-rNSE	&Spatial+Scale	&\textbf{0.9328}	&0.9582	&\textbf{0.9802}	&0.9026	 &\textbf{0.8899}	&\textbf{0.9434}\\
    SOS-H-SSIM	&Spatial+Scale	&0.906	&0.9415	&0.9711	&0.8754	&0.8679	&0.9212\\
  \midrule
    PSNR	&$\setminus$	&0.9081	&0.8923	&0.984	&0.8111	&0.8941	&0.8839\\
    SSIM	&$\setminus$	&0.9606	&0.9739	&0.9693	&0.9515	&0.9551	&0.9481\\
  \bottomrule
  \end{tabular}
  \label{Table_3}
\end{table*}

\begin{figure}[!tb]
    \centering
    \includegraphics[width=0.45\textwidth]{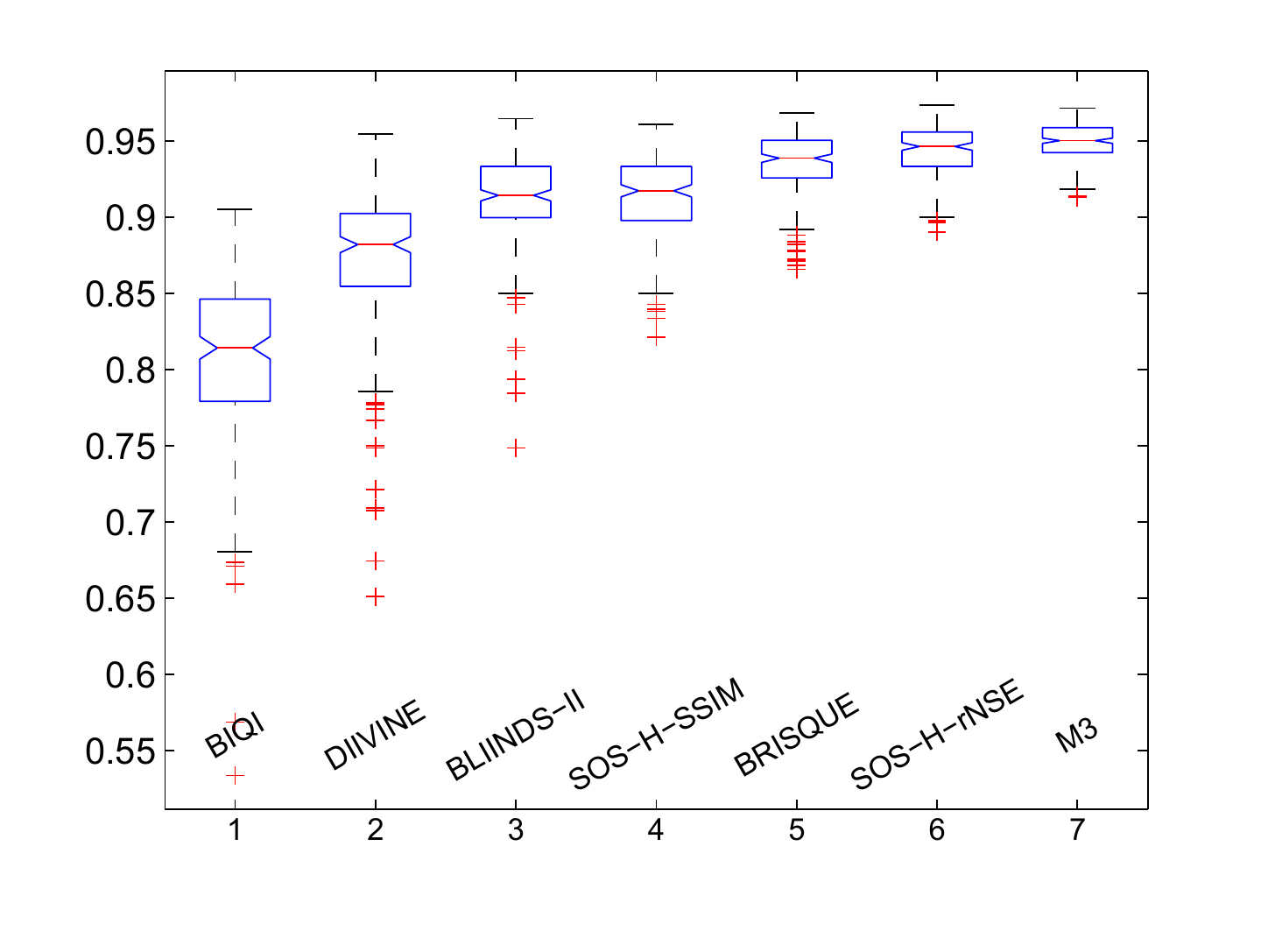}
    \vspace{-5mm}
    \caption{Box plot of the BIQA methods' performance on LIVE database in terms of SRC.}
    \label{fig_boxplot}
    \vspace{-5mm}
\end{figure}

\begin{table*}
    \caption{\textsc{P-values of t-test between each pair of BIQA methods.}}
    \centering
    \begin{tabular}{c|*7{c}}
    \toprule
	       &BIQI	&DIIVINE	&BLIINDS-II	&BRISQUE	&SOS-H-SSIM	&SOS-H-rNSE	&M3\\
    \midrule
    BIQI	&$\setminus$	&1	&1	&1	&1	&1	&1\\
    DIIVINE	&0	&$\setminus$	&1	&1	&1	&1	&1\\
    BLIINDS-II	&0	&0	&$\setminus$	&1	&0.79	&1	&1\\
    BRISQUE	&0	&0	&0	&$\setminus$	&0	&1	&1\\
    SOS-H-SSIM	&0	&0	&0.21	&1	&$\setminus$	&1	&1\\
    SOS-H-rNSE	&0	&0	&0	&0	&0	&$\setminus$	&1\\
    M3	&0	&0	&0	&0	&0	&0	&$\setminus$ \\
    \bottomrule
    \end{tabular}

    \label{table_pvalue}
\end{table*}

\subsection{Performance Comparison with Existing BIQA Methods}
In TABLE~\ref{Table_3}, we compare the performances of the proposed SOS-based methods with existing state-of-the-art BIQA methods, including BIQI~\cite{moorthy2010biqi_9}, BLIINDS-II~\cite{Saad2011model_13}, DIIVINE~\cite{Moorthy2011blind_23}, GRNN~\cite{li2011blind_26}, visual codebook based method (CBIQ)~\cite{ye2012no}, local dependency based method (LD-GS and LD-TS)~\cite{gao2011universal_27}, BRISQUE~\cite{mittal2012no_6} and M3~\cite{xue2014blind_7}. The results of these competitors are either sourced from their original publications or computed by using the source codes provided by the authors. The results of the classical PSNR and SSIM indices are also presented for reference. To save space, only the result of SRC index is shown in TABLE~\ref{Table_3}. The top three results are highlighted with boldface for each column.

When the entire LIVE database is considered, the proposed two SOS-H methods show very competitive performance with the state-of-the-art BIQA methods. The top two methods on LIVE database are M3, SOS-H-rNSE and BRISUQE. SOS-H-SSIM beats all the wavelet-based methods. Box plots of the results on LIVE database are presented in Fig.~\ref{fig_boxplot} for a more intuitive comparison. To investigate the significance of difference between the performances of these BIQA methods, the right-tailed t-test with a significance level of 0.01 is conducted for each pair of BIQA methods. The null hypothesis is that the mean of the SRC values of the two methods are equal. The alternative hypothesis is that the mean SRC value of the method in the row is greater than that of the method in the column. The resulting \emph{p}-values of the tests are shown in TABLE~\ref{table_pvalue}, and a small \emph{p}-value favors the alternative hypothesis. Again, we can see that the proposed SOS-H-rNSE delivers excellent performance, and it is only beaten by M3.

\begin{table*}
    \caption{\textsc{The database independency of SOS-H. The models are trained on LIVE and applied on CSIQ and TID2013. Only the four distortions common to all the three databases are considered.}}
    \centering
    \begin{tabular}{c|*6{c}}
    \toprule
	        &DIIVINE	&BLIINDS-II	&BRISQUE	&M3	&SOS-H-SSIM	&SOS-H-rNSE\\
    \midrule
    CSIQ	&0.857	&0.888	&0.899	&0.911	&0.898	&0.907\\
    TID2013	&0.860	&0.895	&0.891	&0.923	&0.897	&0.913\\
    \bottomrule
    \end{tabular}

    \label{table_crossdatabase}
\end{table*}

\begin{table*}
    \caption{\textsc{The performance of SOS-H with more distortion types in the TID2013 database. }}
    \centering

    \begin{tabular}{c|*6{c}}
    \toprule
SRC	&SOS-H-rNSE	&SOS-H-SSIM	&BRISQUE	&BLIINDS-II	&DIIVINE	&M3\\
    \midrule
Additive Gaussian noise	&\textbf{0.821}	&0.774	&\textbf{0.778}	&0.628	&0.709	&0.769\\
Additive noise more in color	&0.501	&0.542	&\textbf{0.554}	&0.357	&0.431	 &\textbf{0.583}\\
Spatially correlated noise	&0.728	&0.761	&\textbf{0.830}	&0.689	&\textbf{0.816}	 &0.783\\
\rowcolor{gray!50}
Masked noise	&0.261	&\textbf{0.307}	&0.172	&0.281	&0.111	&\textbf{0.504}\\
High frequency noise	&0.876	&\textbf{0.893}	&0.855	&0.772	&0.816	&\textbf{0.884}\\
Impulse noise	&0.739	&0.698	&\textbf{0.815}	&0.607	&\textbf{0.789}	&0.718\\
Quantization noise	&0.579	&\textbf{0.748}	&0.695	&0.639	&0.535	&0.819\\
Gaussian blur	&0.863	&0.826	&0.856	&0.855	&\textbf{0.915}	&\textbf{0.872}\\
Image denoising	&\textbf{0.803}	&0.693	&0.551	&\textbf{0.797}	&0.723	&0.771\\
JPEG compression	&\textbf{0.832}	&0.748	&0.756	&0.706	&0.725	&\textbf{0.819}\\
JPEG2000 compression	&\textbf{0.898}	&0.793	&0.780	&0.850	&0.861	&\textbf{0.873}\\
\rowcolor{gray!50}
JPEG transmission errors	&\textbf{0.435}	&0.165	&0.231	&0.409	&0.343	 &\textbf{0.423}\\
JPEG2000 transmission errors	&0.565	&0.633	&0.695	&0.696	&\textbf{0.717}	 &\textbf{0.723}\\
\rowcolor{gray!50}
Non eccentricity pattern noise	&\textbf{0.182}	&0.131	&0.126	&0.176	&0.134	 &\textbf{0.212}\\
\rowcolor{gray!50}
Local block-wise distortions	&0.146	&0.206	&0.203	&\textbf{0.290}	&\textbf{0.298}	 &0.280\\
\rowcolor{gray!50}
Mean shift	&0.127	&\textbf{0.217}	&0.112	&0.185	&\textbf{0.197}	&0.090\\
\rowcolor{gray!50}
Contrast change	&0.161	&0.056	&0.058	&0.085	&\textbf{0.347}	&\textbf{0.301}\\
\rowcolor{gray!50}
Change of color saturation	&0.099	&0.175	&0.092	&0.022	&\textbf{0.213}	 &\textbf{0.185}\\
Multiplicative Gaussian noise	&0.695	&\textbf{0.720}	&0.621	&0.626	&0.666	 &\textbf{0.709}\\
\rowcolor{gray!50}
Comfort noise	&0.142	&0.021	&0.165	&0.084	&\textbf{0.265}	&\textbf{0.229}\\
Lossy compression of noisy images	&0.628	&0.639	&0.531	&0.454	&\textbf{0.677}	 &\textbf{0.704}\\
Image color quantization with dither	&\textbf{0.837}	&0.815	&0.827	&0.789	&0.802	 &\textbf{0.858}\\
Chromatic aberrations	&0.678	&0.707	&\textbf{0.731}	&0.596	&\textbf{0.775}	&0.644\\
Sparse sampling and reconstruction	&0.847	&0.819	&0.807	&\textbf{0.861}	&0.843	 &\textbf{0.922}\\
ALL	&\textbf{0.608}	&0.569	&0.558	&0.576	&0.593	&\textbf{0.687}\\
    \bottomrule
    \end{tabular}
    \label{table_tid2013}
\end{table*}

When each distortion type is considered, SOS-H-rNSE shows top performance on JP2K, WN, and FF, while SOS-H-SSIM gives inferior SRC value. More specifically, on JP2K images, all methods that based on wavelet features~\cite{moorthy2010biqi_9,Moorthy2011blind_23} fail to give a high SRC value, while the DCT based BLIINDS-II, the gradient and LOG based M3~\cite{xue2014blind_7} and the purposed SOS-H methods show excellent performance. On JPEG images, the two SOS-H methods perform better than the wavelet based methods. On WN images, all the methods based on spatial features show clear advantage over the wavelet, Gabor and DCT based methods. This is due to the fact that pixel based representation in spatial domain is more appropriate for additive noise. On GB images, BRISQUE, M3 and CBIQ-II give the best performance, while the proposed SOS-H methods are still better than the wavelet-based methods. On FF images, it is hard to capture the intrinsic characteristic for quality prediction because FF simultaneously introduces structure shifting, blurring, ringing and color contamination. Among the competing methods, CBIQ-II behaves the best, followed by SOS-H-rNSE and BRISQUE. SOS-H-SSIM only leads to an acceptable performance.

When compared to the FR methods PSNR and SSIM, the two SOS-H methods show an obvious advantage over PSNR, and inferior to the SSIM index.

\subsection{Database Indenpendency}
We examine the database independency of the proposed SOS-H methods as follows: we train a quality prediction model with the SOS-H features on LIVE database and then test the model on CSIQ and TID2013. Note that for CSIQ and TID2013, only images with the four common distortions to LIVE are considered. The results are shown in TABLE~\ref{table_crossdatabase}. Obviously, the two SOS-H methods show good independency of databases. When the LIVE-trained models are tested on CSIQ and TID2013, SOS-H-rNSE gives SRC values competitive with M3, and SOS-H-SSIM works on par with BRISQUE. DIIVINE shows less stable performance in this case.

\subsection{More Distortion Types}
We further test the performance of the proposed SOS-H methods with more distortions by using the TID2013 database, which has a wide range of distortions. $80\%$ of the images are used for training and the rest $20\%$ for testing. The procedure is the same as described in subsection~\ref{III.B}. The obtained median SRC values are presented in TABLE~\ref{table_tid2013}.  The top two methods in each row are highlighted in bold for each distortion. (The detailed explanations of the distortions can be found in~\cite{tid2013_22}.) As can be seen, SOS-H-rNSE shows better performance than other BIQA methods, except for M3, on the entire TID2013 database. On some distortions, all BIQA methods fail to give acceptable performance. We shade the rows in TABLE~\ref{table_tid2013} where all methods obtain SRC values less than 0.5. Examples of these distortions include JPEG transmission errors, non-eccentricity pattern noise, block-wise distortion with different intensity, mean shift, contrast change, color saturation change, and comfort noise. The failure in these cases may be due to the fact that all the current BIQA methods make use of structure features which are not capable of capturing the non-structural distortions, such as color aberration, mean shift, etc.

\subsection{Similarity Function of MSE}
To further validate the effectiveness of the proposed SOS framework, we take MSE as the similarity function in SOS:
\begin{equation}\label{EQ_MSE}
    LSM_{x,y}(\textbf{I},\textbf{R})=\frac{log(1+|\textbf{I}(x,y)-\textbf{R}(x,y)|^2)}{10}.
\end{equation}
Note that we take a logarithm transform of MSE in order to better compute the histogram of LSM. The features for quality prediction are extracted in the same way as that in SOS-H-rNSE and SOS-H-SSIM. The resulting SOS-based method is denoted as SOS-H-MSE.

\begin{table}

  \centering
  \caption{\textsc{SRC Performance comparison of SOS-H with rNSE, SSIM and MSE as the similarity functions.}}
  \begin{tabular}{c|c|ccc}
  \toprule
  \multicolumn{2}{c|}{Database}& SOS-H-rMSE &SOS-H-SSIM &SOS-H-MSE\\
  \midrule
  \multirow{6}*{LIVE}
    &ALL	&0.943	&0.921	&0.944\\
    &JP2K	&0.933	&0.906	&0.946\\
    &JPEG	&0.958	&0.942	&0.959\\
    &WN	&0.980	&0.971	&0.981\\
    &GB	&0.903	&0.875	&0.928\\
    &FF	&0.890	&0.868	&0.865\\
  \midrule
  \multirow{5}*{CSIQ}
    &ALL	&0.910	&0.901	&0.902\\
    &WN	&0.923	&0.920	&0.917\\
    &JPEG	&0.906	&0.915	&0.918\\
    &JP2K	&0.915	&0.907	&0.905\\
    &GB	&0.902	&0.905	&0.895\\
  \midrule
  \multirow{5}*{TID2013}
    &ALL	&0.928	&0.920	&0.919\\
    &WN	&0.923	&0.914	&0.918\\
    &GB	&0.913	&0.914	&0.915\\
    &JPEG	&0.906	&0.901	&0.893\\
    &JP2K	&0.931	&0.923	&0.923\\
  \bottomrule
  \end{tabular}
  \label{Table_mse}
\end{table}

TABLE~\ref{Table_mse} compares the performances of the three SOS-H based methods. Under the framework of SOS, MSE gives similar performance to SSIM on databases of CSIQ and TID2013, and the same performance as rNSE on LIVE. We can draw that the effectiveness of SOS framework can be demonstrated by all the three similarity functions of rNSE, SSIM and MSE. The MSE-based SOS-H even shows slightly better performance in terms of SRC on the LIVE database. For each distortion type, SOS-H-rNSE and SOS-H-MSE show similar results. The good results of SOS-H-MSE can be explained as follows. MSE computes the squared difference between the original image and its shifted or smoothed version. This is similar to the computation of image gradient, which has been shown very effective for image quality assessment~\cite{xue2014gradient_30}. Better performance may be achieved by other potential similarity functions under the SOS framework.

\subsection{Discussions}
It was found that the function of ganglion and lateral geniculate nucleus (LGN) neurons can be modeled by principal component analysis (PCA) based whitening, while the role of PCA is similar to DCT for natural images~\cite{ahmed1974discrete_28}. The responses of simple cells in the primary visual cortex (V1) are similar to the WT outputs and approach to the independent components of natural images~\cite{fischer2007self_29}. However, these transforms may not be effective and efficient to represent distorted images in the context of IQA. The scatter plots in the first two rows of Fig.~\ref{fig_nca} show clearly that the features extracted from DCT and WT domains cannot distinguish well the distorted image and their reference counterpart, and their 2D scatter plots show low relevance to the subjective quality of image.

This fact motivated us to find a different method for NSS calculation. Instead of transformation, we directly compute the intra-scale and inter-scale redundancy in the spatial domain. Interestingly, as shown in the third row of Fig.~\ref{fig_2}, the proposed SOS features can distinguish better the original natural images from their distorted counterparts and the scatter plots show better relevance with subjective quality. Our experiments in the previous sections also validated that the SOS features can predict the perceptual quality very well. Besides, the proposed SOS features works robustly with no strict restriction on the similarity functions. Whether or not our results imply a new physical model of HVS to sense the image quality will be an interesting problem open to investigate, whereas this is out the scope of this paper.

\section{Conclusion}\label{V}
It is well-known that a proper presentation will make the task of image processing more easily, so does for the task of image quality assessment. In this paper, we proposed a new feature representation framework which aims to capture the statistics of self-similarity (SOS) for natural images. Different from previous methods, SOS directly measures the redundancy existing in an image, rather than describing the structure in a redundancy reduced domain. The computed local similarity map (LSM) can portray the local correlation across space and scales, both of which will be altered by image distortion. The statistics of these LSMs, i.e., the SOS features, were validated to be able to more effectively capture the distortion degree than previous features that are based on image decomposition. Especially, when the LSM histogram features are utilized, very competitive performance can be achieved on the benchmark databases. New similarity functions can be introduced or designed under the framework of SOS for better performance in the future study.


\ifCLASSOPTIONcaptionsoff
  \newpage
\fi



\bibliographystyle{IEEEtran}
\bibliography{ref_list}

\begin{thebibliography}{10}
\providecommand{\url}[1]{#1}
\csname url@samestyle\endcsname
\providecommand{\newblock}{\relax}
\providecommand{\bibinfo}[2]{#2}
\providecommand{\BIBentrySTDinterwordspacing}{\spaceskip=0pt\relax}
\providecommand{\BIBentryALTinterwordstretchfactor}{4}
\providecommand{\BIBentryALTinterwordspacing}{\spaceskip=\fontdimen2\font plus
\BIBentryALTinterwordstretchfactor\fontdimen3\font minus
  \fontdimen4\font\relax}
\providecommand{\BIBforeignlanguage}[2]{{%
\expandafter\ifx\csname l@#1\endcsname\relax
\typeout{** WARNING: IEEEtran.bst: No hyphenation pattern has been}%
\typeout{** loaded for the language `#1'. Using the pattern for}%
\typeout{** the default language instead.}%
\else
\language=\csname l@#1\endcsname
\fi
#2}}
\providecommand{\BIBdecl}{\relax}
\BIBdecl

\bibitem{wang2006modern_1}
Z.~Wang and A.~C. Bovik, ``Modern image quality assessment,'' \emph{Synthesis
  Lectures on Image, Video, and Multimedia Processing}, vol.~2, no.~1, pp.
  1--156, 2006.

\bibitem{zhang2011nser_16}
M.~Zhang, X.~Mou, and L.~Zhang, ``Non-shift edge based ratio (nser): An image
  quality assessment metric based on early vision features,'' \emph{Signal
  Processing Letters, IEEE}, no.~99, pp. 1--1, 2011.

\bibitem{larson2010most_21}
E.~C. Larson and D.~M. Chandler, ``Most apparent distortion: full-reference
  image quality assessment and the role of strategy,'' \emph{Journal of
  Electronic Imaging}, vol.~19, no.~1, pp. 011\,006--011\,006, 2010.

\bibitem{xue2014gradient_30}
W.~Xue, L.~Zhang, X.~Mou, and A.~C. Bovik, ``Gradient magnitude similarity
  deviation: a highly efficient perceptual image quality index,'' \emph{Image
  Processing, IEEE Transactions on}, vol.~23, no.~2, pp. 684--695, 2014.

\bibitem{SongnanFR}
S.~Li, F.~Zhang, L.~Ma, and K.~N. Ngan, ``Image quality assessment by
  separately evaluating detail losses and additive impairments,''
  \emph{Multimedia, IEEE Transactions on}, vol.~13, no.~5, pp. 935--949, Oct
  2011.

\bibitem{zhang2011fsim}
L.~Zhang, X.~Mou, and D.~Zhang, ``Fsim: A feature similarity index for image
  quality assessment,'' \emph{Image Processing, IEEE Transactions on}, no.~99,
  pp. 1--1, 2011.

\bibitem{li2009reduced}
Q.~Li and Z.~Wang, ``Reduced-reference image quality assessment using divisive
  normalization-based image representation,'' \emph{Selected Topics in Signal
  Processing, IEEE Journal of}, vol.~3, no.~2, pp. 202--211, 2009.

\bibitem{xue2010weibull}
W.~Xue and X.~Mou, ``Reduced reference image quality assessment based on
  weibull statistics,'' in \emph{Quality of Multimedia Experience (QoMEX), 2010
  Second International Workshop on}.\hskip 1em plus 0.5em minus 0.4em\relax
  IEEE, 2010, pp. 1--6.

\bibitem{mou2012reduced}
X.~Mou, W.~Xue, and L.~Zhang, ``Reduced reference image quality assessment via
  sub-image similarity based redundancy measurement,'' \emph{Proceedings of
  SPIE}, vol. 8291, p. 82911S, 2012.

\bibitem{LinMA_DCT_RR}
L.~Ma, S.~Li, F.~Zhang, and K.~N. Ngan, ``Reduced-reference image quality
  assessment using reorganized dct-based image representation,''
  \emph{Multimedia, IEEE Transactions on}, vol.~13, no.~4, pp. 824--829, Aug
  2011.

\bibitem{VIF_RR}
J.~Wu, W.~Lin, G.~Shi, and A.~Liu, ``Reduced-reference image quality assessment
  with visual information fidelity,'' \emph{Multimedia, IEEE Transactions on},
  vol.~15, no.~7, pp. 1700--1705, Nov 2013.

\bibitem{shahid2014no_2}
M.~Shahid, A.~Rossholm, B.~L{\"o}vstr{\"o}m, and H.-J. Zepernick,
  ``No-reference image and video quality assessment: a classification and
  review of recent approaches,'' \emph{EURASIP Journal on Image and Video
  Processing}, vol. 2014, no.~1, pp. 1--32, 2014.

\bibitem{manap2015non_3}
R.~A. Manap and L.~Shao, ``Non-distortion-specific no-reference image quality
  assessment: A survey,'' \emph{Information Sciences}, vol. 301, pp. 141--160,
  2015.

\bibitem{moorthy2010biqi_9}
A.~Moorthy and A.~Bovik, ``A two-step framework for constructing blind image
  quality indices,'' \emph{Signal Processing Letters, IEEE}, vol.~17, no.~5,
  pp. 513--516, 2010.

\bibitem{tang2011lbiq_10}
H.~Tang, N.~Joshi, and A.~Kapoor, ``Learning a blind measure of perceptual
  image quality,'' in \emph{Computer Vision and Pattern Recognition (CVPR),
  2011 IEEE Conference on}.\hskip 1em plus 0.5em minus 0.4em\relax IEEE, 2011,
  pp. 305--312.

\bibitem{Moorthy2011blind_23}
A.~Moorthy and A.~Bovik, ``Blind image quality assessment: From natural scene
  statistics to perceptual quality,'' \emph{Image Processing, IEEE Transactions
  on}, vol.~20, no.~12, pp. 3350 --3364, dec. 2011.

\bibitem{lu2010no_11}
W.~Lu, K.~Zeng, D.~Tao, Y.~Yuan, and X.~Gao, ``No-reference image quality
  assessment in contourlet domain,'' \emph{Neurocomputing}, vol.~73, no.~4, pp.
  784--794, 2010.

\bibitem{saad2010blinddct_12}
M.~Saad, A.~Bovik, and C.~Charrier, ``A dct statistics-based blind image
  quality index,'' \emph{Signal Processing Letters, IEEE}, vol.~17, no.~6, pp.
  583--586, 2010.

\bibitem{Saad2011model_13}
M.~A. Saad, A.~C. Bovik, and C.~Charrier, ``Blind image quality assessment: A
  natural scene statistics approach in the dct domain,'' \emph{Image
  Processing, IEEE Transactions on}, vol.~21, no.~8, pp. 3339--3352, 2012.

\bibitem{teo1994perceptual_4}
P.~Teo and D.~Heeger, ``Perceptual image distortion,'' in \emph{Image
  Processing, 1994. Proceedings. ICIP-94., IEEE International Conference},
  vol.~2.\hskip 1em plus 0.5em minus 0.4em\relax IEEE, 1994, pp. 982--986.

\bibitem{ruderman1994statistics_5}
D.~L. Ruderman and W.~Bialek, ``Statistics of natural images: Scaling in the
  woods,'' \emph{Physical review letters}, vol.~73, no.~6, p. 814, 1994.

\bibitem{mittal2012no_6}
A.~Mittal, A.~K. Moorthy, and A.~C. Bovik, ``No-reference image quality
  assessment in the spatial domain,'' \emph{Image Processing, IEEE Transactions
  on}, vol.~21, no.~12, pp. 4695--4708, 2012.

\bibitem{xue2014blind_7}
W.~Xue, X.~Mou, L.~Zhang, A.~C. Bovik, and X.~Feng, ``Blind image quality
  assessment using joint statistics of gradient magnitude and laplacian
  features,'' \emph{Image Processing, IEEE Transactions on}, vol.~23, no.~11,
  pp. 4850--4862, 2014.

\bibitem{attneave1954some_8}
F.~Attneave, ``Some informational aspects of visual perception.''
  \emph{Psychological review}, vol.~61, no.~3, p. 183, 1954.

\bibitem{kersten1987predictability}
D.~Kersten, ``Predictability and redundancy of natural images,'' \emph{JOSA A},
  vol.~4, no.~12, pp. 2395--2400, 1987.

\bibitem{dam2003front_14}
E.~Dam and B.~ter Haar~Romeny, ``Front end vision and multi-scale image
  analysis,'' \emph{Deep Structure I, II \& III}, no. 1-4020, pp. 1507--0,
  2003.

\bibitem{wang2004ssim_15}
Z.~Wang, A.~Bovik, H.~Sheikh, and E.~Simoncelli, ``Image quality assessment:
  From error visibility to structural similarity,'' \emph{Image Processing,
  IEEE Transactions on}, vol.~13, no.~4, pp. 600--612, 2004.

\bibitem{xue2011rnse_17}
W.~Xue and X.~Mou, ``An image quality assessment metric based on non-shift
  edge,'' in \emph{Image Processing (ICIP), 2011 18th IEEE International
  Conference on}.\hskip 1em plus 0.5em minus 0.4em\relax IEEE, 2011, pp.
  3309--3312.

\bibitem{smola2004svrtutorial_18}
A.~Smola and B.~Sch{\"o}lkopf, ``A tutorial on support vector regression,''
  \emph{Statistics and computing}, vol.~14, no.~3, pp. 199--222, 2004.

\bibitem{roweis17neighbourhood_19}
S.~Roweis, G.~Hinton, and R.~Salakhutdinov, ``Neighbourhood component
  analysis,'' in \emph{Neural Information Processing Systems}, vol.~17, pp.
  513--520.

\bibitem{sheikh2007live_20}
H.~Sheikh, Z.~Wang, L.~Cormack, and A.~Bovik, ``Live image quality assessment
  database release 2 (2005).''

\bibitem{tid2013_22}
N.~Ponomarenko, O.~Ieremeiev, V.~Lukin, K.~Egiazarian, L.~Lin, J.~Astola,
  B.~Vozel, K.~Chehdi, M.~Carli, F.~Battisti, and C.-C. Jay~Kuo, ``Color image
  database tid2013: Peculiarities and preliminary results,'' \emph{Advances of
  Modern Radioelectronics}, vol.~10, no.~10, pp. 30--45, 2009.

\bibitem{video2003final_24}
VQEG, ``Final report from the video quality experts group on the validation of
  objective models of video quality assessment, phase ii,'' \emph{VQEG, Aug},
  2003.

\bibitem{chang2011libsvm_25}
C.~Chang and C.~Lin, ``Libsvm: a library for support vector machines,''
  \emph{ACM Transactions on Intelligent Systems and Technology (TIST)}, vol.~2,
  no.~3, p.~27, 2011.

\bibitem{li2011blind_26}
C.~Li, A.~Bovik, and X.~Wu, ``Blind image quality assessment using a general
  regression neural network,'' \emph{Neural Networks, IEEE Transactions on},
  vol.~22, no.~5, pp. 793--799, 2011.

\bibitem{gao2011universal_27}
F.~Gao, X.~Gao, D.~Tao, X.~Li, L.~He, and W.~Lu, ``Universal no reference image
  quality assessment metrics based on local dependency,'' in \emph{Pattern
  Recognition (ACPR), 2011 First Asian Conference on}.\hskip 1em plus 0.5em
  minus 0.4em\relax IEEE, 2011, pp. 298--302.

\bibitem{ye2012no}
P.~Ye and D.~Doermann, ``No-reference image quality assessment using visual
  codebooks,'' \emph{Image Processing, IEEE Transactions on}, vol.~21, no.~7,
  pp. 3129--3138, 2012.

\bibitem{ahmed1974discrete_28}
N.~Ahmed, T.~Natarajan, and K.~R. Rao, ``Discrete cosine transform,''
  \emph{Computers, IEEE Transactions on}, vol. 100, no.~1, pp. 90--93, 1974.

\bibitem{fischer2007self_29}
S.~Fischer, F.~{\v{S}}roubek, L.~Perrinet, R.~Redondo, and G.~Crist{\'o}bal,
  ``Self-invertible 2d log-gabor wavelets,'' \emph{International Journal of
  Computer Vision}, vol.~75, no.~2, pp. 231--246, 2007.

\end{thebibliography}
%

%


\begin{IEEEbiographynophoto}{Wufeng Xue}
 received the B.Sc. degree in automatic engineering from the School of Electronic and Information Engineering, Xi'an Jiaotong University, Xi'an, China, in 2009. He is currently pursuing the Ph.D. degree with the Institute of Image Processing and Pattern Recognition, Xi'an Jiaotong University. His research interest focuses on perceptual quality of visual signals.
\end{IEEEbiographynophoto}


\begin{IEEEbiographynophoto}{Xuanqin Mou}
(M'08) has been with the Institute of Image Processing and Pattern Recognition (IPPR), Electronic and Information Engineering School, Xi'an Jiaotong University, since 1987. He has been an Associate Professor since 1997, and a Professor since 2002. He is currently the Director of IPPR. Dr. Mou served as the member of the 12th Expert Evaluation Committee for the National Natural Science Foundation of China, the Member of the 5th and 6th Executive Committee of China Society of Image and Graphics, the Vice President of Shaanxi Image and Graphics Association. He has authored or co-authored more than 200 peer-reviewed journal or conference papers. He has been granted as the Yung Wing Award for Excellence in Education, the KC Wong Education Award, the Technology Academy Award for Invention by the Ministry of Education of China, and the Technology Academy Awards from the Government of Shaanxi Province, China.
\end{IEEEbiographynophoto}

\begin{IEEEbiographynophoto}{Lei Zhang}
 (M'04, SM'14) received the B.Sc. degree in 1995 from Shenyang Institute of Aeronautical Engineering, Shenyang, P.R. China, the M.Sc. and Ph.D degrees in Control Theory and Engineering from Northwestern Polytechnical University, Xi'an, P.R. China, respectively in 1998 and 2001. From 2001 to 2002, he was a research associate in the Dept. of Computing, The Hong Kong Polytechnic University. From Jan. 2003 to Jan. 2006 he worked as a Postdoctoral Fellow in the Dept. of Electrical and Computer Engineering, McMaster University, Canada. In 2006, he joined the Dept. of Computing, The Hong Kong Polytechnic University, as an Assistant Professor. Since Sept. 2010, he has been an Associate Professor in the same department. His research interests include Image and Video Processing, Computer Vision, Pattern Recognition and Biometrics, etc. Dr. Zhang has published about 200 papers in those areas. Dr. Zhang is currently an Associate Editor of IEEE Trans. on CSVT and Image and Vision Computing. He was awarded the 2012-13 Faculty Award in Research and Scholarly Activities. More information can be found in his homepage http://www4.comp.polyu.edu.hk/~cslzhang/.
\end{IEEEbiographynophoto}




\end{document}